\documentclass{article}

\usepackage{arxiv}

\renewcommand{\shorttitle}{app.build: Environment Scaffolding for Agentic Code Generation}

\usepackage[utf8]{inputenc}
\usepackage[T1]{fontenc}
\usepackage{hyperref}
\usepackage{url}
\usepackage{xurl}  
\usepackage{amsmath,amssymb,amsfonts}
\usepackage{graphicx}
\usepackage{textcomp}
\usepackage{booktabs}
\usepackage{threeparttable}
\usepackage{float}
\usepackage{natbib}

\title{app.build: A Production Framework for Scaling Agentic Prompt-to-App Generation with Environment Scaffolding\thanks{Accepted to SANER 2026 Industrial Track. Code: \url{https://github.com/neondatabase/appdotbuild-agent}}}

\author{
  Evgenii Kniazev$^{1,\dagger}$ \and
  Arseny Kravchenko$^{1,\dagger}$ \and
  Igor Rekun$^{1,\dagger}$ \and
  James Broadhead$^{1}$ \and
  Nikita Shamgunov$^{1}$ \and
  Pranav Sah$^{2}$ \and
  Pratik Nichite$^{2}$ \and
  Ivan Yamshchikov$^{2}$ \\
  \\
  $^{1}$Databricks \\
  $^{2}$THWS University of Applied Sciences W\"urzburg-Schweinfurt (CAIRO) \\
  $^{\dagger}$Equal contribution \\
  \\
  \texttt{eng-appbuild@databricks.com}
}

\date{}

\begin{document}
\maketitle

\begin{abstract}
Engineering teams increasingly experiment with LLM agents to synthesize full-stack web applications, yet production reliability and code generation reproducibility remain the blocking issues. Ongoing improvements of foundational models alone do not reliably translate into deployable software; what matters in practice is the environment that constrains, validates, and repairs model outputs.

We present the app.build framework and report our industrial experience using environment scaffolding (stack-aware generate$\rightarrow$validate$\rightarrow$repair loops, sandboxed execution, and policy gates) to turn prompt-to-app generation into a dependable workflow. We conducted 300 end-to-end generation experiments with automated validation metrics, complemented by detailed human quality assessment on 30 representative prompts. The framework has been deployed in production and generated 3000+ user applications during 4 months of operation.

Across end-to-end app-building tasks, structured validators and code execution isolation improve the rate of viable apps (viability = pass boot + prompt-correspondence smoke checks) to 73.3\% in our tests, while generic end-to-end browser tests introduce brittleness. Large-scale automated metrics (n=300) reveal that open-weights models achieve 80.8\% performance of top closed model at 8.2$\times$ lower cost per viable app, with validation ablations showing lightweight smoke checks and backend contract tests deliver most reliability lift, whereas broad end-to-end suites often reject working apps.

This paper frames the problem as a software engineering challenge (reliability, maintainability, and cost in agentic development), provides a reproducible evaluation protocol validated at production scale, and distills lessons for practitioners deploying LLM agents. We release the open-source framework (650+ stars) and an artifact to reproduce the main tables.
\end{abstract}


\section{Introduction}
\label{sec:introduction}

\subsection{The Production Reliability Gap}

The promise of Large Language Model (LLM) agents for automated software development has attracted significant industrial interest, yet a critical gap persists between research benchmarks and production requirements. While research systems demonstrate impressive capabilities on isolated benchmarks---HumanEval \citep{chen2021evaluating} leaders achieve 90\%+ pass rates on function-level tasks, and LiveCodeBench \citep{jain2024livecodebench} reaches over 80\% on real GitHub issues---these metrics do not translate to deployable software in industrial contexts.

The gap manifests in three critical dimensions that affect practitioner adoption:

\textbf{Reliability under constraints.} Production systems must operate within fixed time and cost budgets while maintaining deterministic quality gates. LLMs generate probabilistically, producing syntactically correct code that fails integration tests, violates security policies, or exhibits subtle runtime defects \citep{liu2023your}.

\textbf{Reproducibility and debugging.} When generation fails, practitioners need actionable diagnostics. Model-centric approaches offer little guidance for iterative refinement. Practitioners need structured validation feedback that pinpoints specific failure modes, so that repairs can be targeted effectively.

\textbf{Economic viability.} At scale, token costs and iteration cycles determine feasibility. Closed frontier models like Claude Sonnet 4 \citep{anthropic2024claude4} achieve high success rates but at significant costs. For teams generating hundreds of applications, these costs compound rapidly. The industry needs cost-performance tradeoffs: where can open-weights models substitute for frontier models? What validation overhead is justified by reliability gains?

We claim that leaning on model-only improvements is insufficient. The prevailing approach treats reliability as a model capability problem---scale parameters, improve training data, refine prompts. However, our production experience generating thousands of applications reveals that environment design matters more than model selection for industrial deployment. A frontier model without validation produces unreliable apps; industry needs explicit tradeoffs between cost, speed, and correctness. Production-ready systems require frameworks that integrate validation, isolation, and repair as first-class concerns---not post-hoc additions to model outputs. Recent surveys \citep{jiang2024survey,paul2024benchmarks} note the field requires a shift from model-centric to environment-centric design.

\subsection{Our Approach: Environment Scaffolding}

\textbf{Definition.} We define \emph{environment scaffolding (ES)} as an \textbf{environment-first} paradigm for LLM-based code generation where the model operates inside a structured sandbox that constrains actions and provides continuous, deterministic feedback. Rather than relying on larger models or prompt-only techniques, ES \emph{improves the context} around the model --- shaping the action space, providing templates and tools, and validating each step --- so that creativity is channeled into \emph{safe, verifiable} outcomes.

\paragraph{How environment scaffolding works in practice}
Environment scaffolding structures LLM-based code generation around four core practices that address production reliability requirements:

\textbf{Structured task decomposition.} Rather than asking the model to generate an entire application at once, we break work into explicit stages (schema $\rightarrow$ Application Programming Interface (API) $\rightarrow$ User Interface (UI)) with defined inputs, outputs, and success criteria. This matches how developers actually build software and makes failures easier to diagnose.

\textbf{Multi-layered validation.} After every generation step, deterministic checks run automatically: linters catch syntax errors, type-checkers verify contracts, unit tests validate logic, and smoke tests ensure the app boots. Failures trigger immediate repair loops before moving forward, preventing error accumulation.

\textbf{Runtime isolation.} Every generation and test runs in an isolated sandbox with ephemeral state. If the model generates code that crashes or corrupts data, the container resets cleanly. This enables aggressive trial-and-error without risk to production systems.

\textbf{Model-agnostic design.} The scaffolding layer sits between your workflow and the LLM, allowing you to swap models (e.g., from Claude to Qwen) without rewriting validation logic. This protects against vendor lock-in and enables cost-performance optimization.

\begin{figure}[t]
  \centering
  \includegraphics[width=\linewidth]{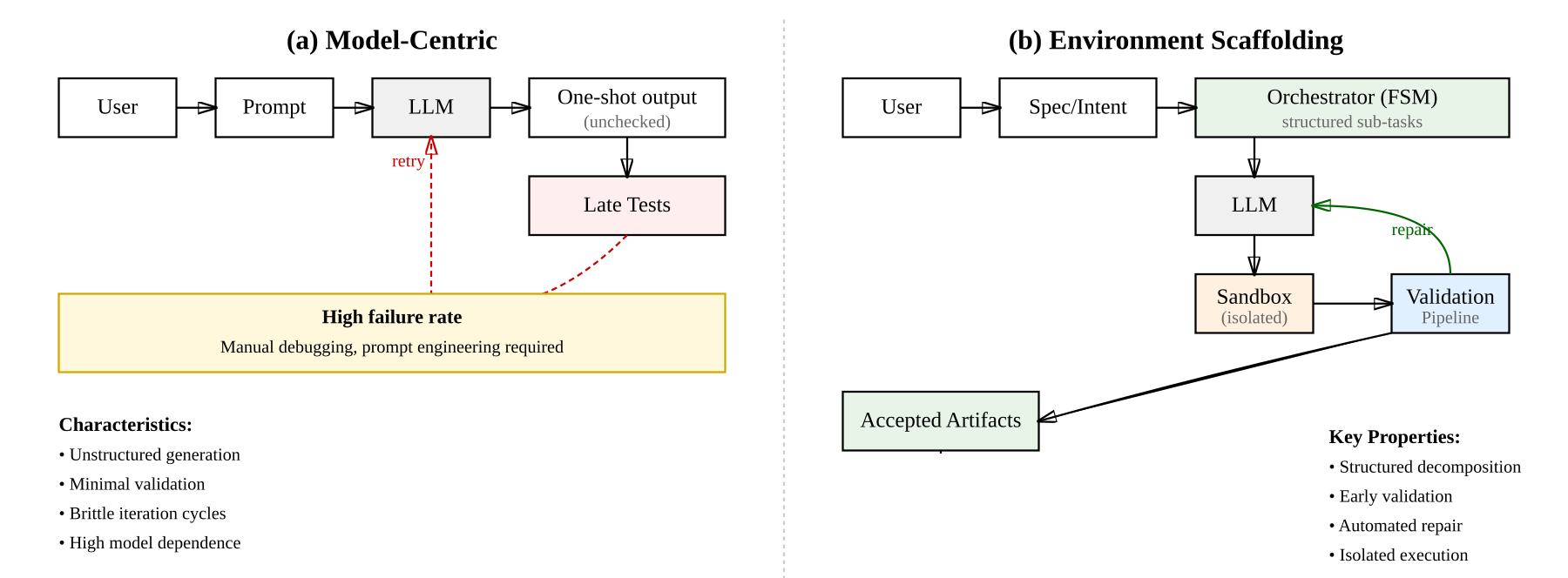}
  \caption{\textbf{Environment scaffolding vs.\ model-centric generation.} ES wraps the model with a finite, validated workflow that catches errors early and repairs them before proceeding.}
  \label{fig:es-vs-model}
\end{figure}

\paragraph{Why this differs from model-centric approaches}
Most existing systems prompt an LLM to generate code and then validate the complete output. This works for simple scripts but fails for full-stack applications where a single integration error can invalidate hours of generation work. Environment scaffolding instead enforces generate$\rightarrow$validate$\rightarrow$repair at each step, catching errors early when they are cheap to fix. Figure~\ref{fig:es-vs-model} and Table~\ref{tab:es-contrast} illustrate this architectural difference.

\begin{table}[t]
\centering
\footnotesize
\begin{threeparttable}
\caption{Environment Scaffolding (ES) vs.\ Model-Centric Generation}
\label{tab:es-contrast}
\begin{tabular}{@{}p{1.8cm}p{2.8cm}p{2.8cm}@{}}
\toprule
\textbf{Aspect} & \textbf{Model-Centric} & \textbf{ES (Ours)} \\
\midrule
Task decomp. & Single/loosely guided; no fixed structure &
Explicit FSM: schema $\rightarrow$ API $\rightarrow$ UI \\
Validation & Late or ad-hoc &
Per-step: linters, types, tests \\
Error recovery & Manual/ad-hoc &
Auto repair loop w/ feedback \\
Execution & Often on host &
Isolated containers \\
Model dep. & Strong (prompt-specific) &
Model-agnostic \\
Observability & Limited logs &
Per-step metrics, artifacts \\
\bottomrule
\end{tabular}
\end{threeparttable}
\end{table}

\subsection{Contributions}

Our work advances \emph{environment-first} agent design. The main contributions are:

\begin{itemize}
  \item \textbf{Environment Scaffolding Paradigm.} We formalize \emph{environment scaffolding (ES)} and show how structuring the action space with per-step validation enables reliable code generation without model-specific tricks.
  \item \textbf{Open-Source Framework (app.build).} We release an implementation of ES that targets three stacks (TypeScript/tRPC, PHP/Laravel, Python/NiceGUI) and ships with validators and deployment hooks. The framework has gained 650+ GitHub stars and 89 forks, demonstrating practitioner adoption.
  \item \textbf{Two-Tier Empirical Evaluation.} We conduct 300 end-to-end generation experiments with automated metrics (success rate, cost, tokens, duration) plus detailed human evaluation on 30 representative prompts with 6-criteria quality rubric. This methodology combines automated consistency checks at scale with manual viability assessment on representative samples.
  \item \textbf{Production-Scale Validation.} The framework has been deployed in production since June 2025, generating over 3000 user applications during the first 4 months with hundreds of applications generated daily at peak usage, providing ecological validity beyond controlled experiments.
  \item \textbf{Cost-Performance Analysis.} We quantify validation overhead through token usage and cost-per-viable-app metrics, showing open-weights models (Qwen3) achieve 70\% success at 8.2x lower cost (\$0.61 vs \$5.01 per viable app), while validation ablations reveal that comprehensive testing increases costs by~\$40 per cohort but catches real defects.
  \item \textbf{Methodological Insight.} We find that improving the \emph{environment} (constraints, tests, repair loops) often matters more than scaling the model for production reliability, with lightweight smoke tests and backend validation providing most gains while End-to-End (E2E) browser tests introduce brittleness.
\end{itemize}

\section{Related Work}
\label{sec:related-work}

\textbf{Repository-level agentic SE (2024-2025).} The evolution of AI coding agents has progressed from code completion to autonomous software engineering systems. \textbf{SWE-bench} \citep{jimenez2024swe} established the evaluation standard with 2,294 real GitHub issues from 12 Python projects. Recent agents demonstrate that environment design rivals model capability: \textbf{OpenHands} \citep{wang2024openhands}, published at ICLR 2025, achieves 53\% on SWE-bench Verified through an open platform for generalist agents with agent-computer interfaces. \textbf{SWE-agent} \citep{yang2024swe} showed 12.5\% pass@1 through careful interface design rather than model improvements. Contemporary 2024 agents include \textbf{AutoCodeRover} \citep{zhang2024autocoder}, which combines LLMs with spectrum-based fault localization (19\% on SWE-bench, \$0.43 per issue), and \textbf{Agentless} \citep{xia2024agentless}, challenging architectural complexity with a simple three-phase process (localization, repair, validation) achieving 32\% on SWE-bench Lite.

\textbf{Validation and environment scaffolding.} Production-ready code generation requires validation beyond correctness testing. While early explorations in this space focused on code change classification \citep{kniazev2008automated}, modern frameworks now integrate validation at multiple layers. Test-driven approaches \citep{pan2024ticoder} achieve 45.97\% absolute improvement in pass@1 through interactive generation with dynamic test feedback. \textbf{AST-based validation} \citep{gong2024astt5} provides structural guarantees, with AST-T5 outperforming CodeT5 by 2--3 points through structure-aware pretraining. Tree search methods \citep{li2025s} demonstrate that scaling compute through iterative refinement and parallel branches can significantly improve success rates. Multi-agent systems \citep{hong2023metagpt} show that role-based collaboration with structured validation outperforms single-agent approaches, achieving 85.9\% pass@1 on HumanEval with 100\% task completion on development tasks. For web application generation, sandboxed execution with database provisioning and browser emulation is essential for isolating and validating complex multi-tier systems.

\section{Industrial Context \& System}
\label{sec:method}

\subsection{Problem Formulation}

LLM-based code generation enables rapid prototyping but often produces code that does not meet production standards. We formalize this as an environment design problem where success depends not just on model capability but on the structured constraints and validation feedback provided by the generation environment.

\begin{figure}[t]
  \centering
  \includegraphics[width=\linewidth]{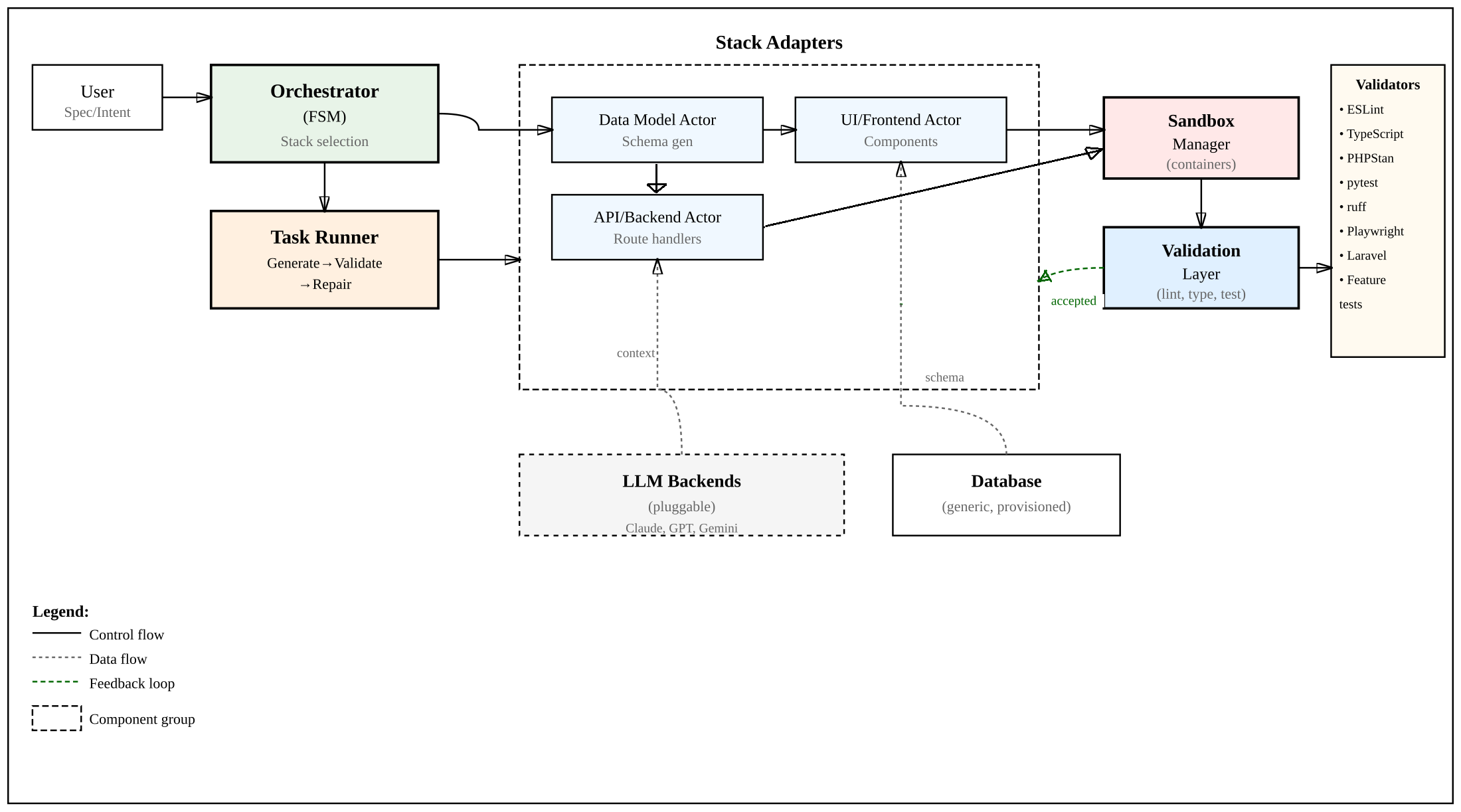}
  \caption{\textbf{app.build architecture} expressed through environment scaffolding. The orchestrator plans stages per stack; each sub-task runs in a sandbox, is validated, and only then merged. Continuous Integration/Continuous Deployment (CI/CD) and database provisioning are integrated.}
  \label{fig:appbuild-arch}
\end{figure}

\subsection{Architecture}

\textbf{High-level design.} The app.build agent implements ES with a central \emph{orchestrator} that decomposes a user's specification into stack-specific stages and executes each stage inside an isolated sandbox with validation before acceptance. The same workflow applies across supported stacks (TypeScript/tRPC, PHP/Laravel, Python/NiceGUI), selected for their deterministic scaffolding patterns and comprehensive validator availability (TypeScript/ESLint/Playwright, PHPStan/Laravel feature tests, pytest/ruff/pyright). Per-stage validators are stack-aware, and the platform provisions managed Postgres databases and CI/CD hooks.

\textbf{Execution loop.} For each sub-task, the agent (i) assembles minimal context (files, interfaces, constraints), (ii) prompts the LLM, (iii) executes the result in a sandbox, (iv) collects validator feedback, and (v) either accepts the artifact or re-prompts to repair. This iterative loop provides robustness without assuming a particular model, and scales by parallelizing sandboxes and caching environment layers.

\section{Experimental Setup}
\label{sec:experimental-setup}

We designed experiments using a custom prompt dataset and metrics to evaluate viability and quality of generated applications.

\subsection{Evaluation Framework}

Our evaluation methodology addresses the core challenge of assessing prompt-to-app generation systems: validating that generated applications not only compile and boot, but actually implement the requested functionality correctly. We combine manual-heavy viability assessment with automated checks across 300 end-to-end experiments, where every generated application undergoes deterministic validation followed by human review on a representative subset.

\textbf{Two-tier evaluation design.} We conduct 300 end-to-end generation runs with automated checks measuring boot success, smoke test passage, cost, token usage, and duration. These automated gates provide fast, deterministic feedback but cannot assess functional correctness or usability. For viability assessment, human evaluators systematically test 30 representative applications using a structured 6-criteria rubric (Section~\ref{sec:scoring}), verifying that generated apps correctly implement requested functionality, handle edge cases, and meet production quality standards. This manual assessment is essential because automated metrics alone cannot detect integration bugs, incorrect business logic, or poor UX that would block production deployment.

\textbf{Viability vs. quality distinction.} We separate binary viability ($V \in \{0,1\}$) from continuous quality ($Q \in [0,10]$). Viability requires only that an application boots successfully and demonstrates basic prompt correspondence---a minimal threshold for deployment consideration. Quality scoring evaluates correctness, completeness, error handling, and code maintainability through systematic human assessment. This distinction reflects industrial practice: automated gates filter non-viable candidates before human review evaluates deployment readiness.

\textbf{Production-scale validation.} Beyond controlled experiments, we validate the framework through production deployment metrics. Since June 2025, the system has generated over 3000 user applications with peak usage exceeding 220 apps/day (Section~\ref{sec:results}). Production data provides ecological validity showing the framework operates reliably with diverse real-world requirements beyond our curated test set.

\textbf{Cost-performance tradeoffs.} All experiments track token consumption (input/output) and API costs to quantify validation overhead. We report both total cost per generation cohort and cost-per-viable-app to reveal true economics: validation may increase upfront costs but reduces the effective cost of successful outcomes by filtering failures early.

\subsection{Prompt Dataset}
\label{sec:prompt-dataset-desc}

The evaluation dataset comprises 30 prompts designed to assess system performance across diverse application development scenarios. Independent human contributors with no prior exposure to the app.build system created evaluation prompts. Contributors developed tasks reflecting authentic development workflows from their professional experience. Prompts were filtered to exclude enterprise integrations, AI/ML compute requirements, or capabilities beyond framework scope. Raw prompts underwent automated post-processing using LLMs to anonymize sensitive information and standardize linguistic structure.
The resulting dataset consists of 30 prompts spanning a complexity spectrum (low: static/single-page UI; medium: single-entity Create, Read, Update, Delete (CRUD); high: multi-entity/custom logic).
See the full list of prompts in Appendix~\ref{sec:prompt-dataset}.

Each application generated by the agent was evaluated by the following metrics, designed to assess its viability and quality under preset time and cost constraints.

\begin{itemize}
\item Viability rate ($V=1$) and non-viability rate ($V=0$)
\item Perfect quality rate ($Q=10$) and quality distribution (mean/median for $V=1$ apps)
\item Validation pass rates by check (AB-01 through AB-06, defined in Section~\ref{sec:scoring})
\item Quality scores ($Q$, 0--10) using the rubric in Section~\ref{sec:scoring}
\item Model/cost comparisons where applicable
\end{itemize}

\subsection{Experimental Configurations}

We designed three experimental configurations to systematically evaluate factors affecting app generation success rates:

\textbf{Configuration 1: Baseline}. We generated baseline tRPC apps with default production setup and all checks ON to assess default generation success rate, cost and time.

\textbf{Configuration 2: Model Architecture Analysis}. Using the tRPC stack, we evaluated open versus closed foundation models. Claude Sonnet 4 served as the baseline coding model, compared against Qwen3-Coder-480B-A35B \citep{qwen2025qwen3} and GPT OSS 120B \citep{openai2025gpt} as open alternatives.

\textbf{Configuration 3: Testing Framework Ablation}. We conducted three ablation studies on the tRPC stack isolating the impact of each type of checks by turning them off independently: (3a) disabled isolated Playwright UI smoke tests; (3b) disabled ESLint checks; and (3c) removed handlers tests, eliminating backend validation.

\subsection{Assessor Protocol and Scoring}
\label{sec:scoring}

To systematically assess generated application quality, we implement a structured evaluation protocol comprising six standardized functional checks executed by human assessors. The evaluation reports two independent outcomes: a binary viability indicator ($V$) and a 0--10 quality score ($Q$). The complete assessor handbook with detailed grading criteria and example graded applications is publicly available~\citep{appbuild2025handbook}.

\textbf{Viability (binary)}:
\begin{equation}
V = \begin{cases}
1 & \text{if AB-01 and AB-02 are not FAIL} \\
0 & \text{otherwise}
\end{cases}
\end{equation}

\textbf{Quality (0--10)}:
\begin{equation}
Q = 10 \times \frac{\sum_{c \in A} w \times s_c}{\sum_{c \in A} w}
\end{equation}

where $A$ is the set of applicable checks (excluding NA); all checks use equal weights prior to NA re-normalization; and per-check grades $s_c$ are mapped as follows:
\begin{itemize}
\item AB-01 (Boot): PASS = 1.0, WARN = 0.5, FAIL = 0.0
\item AB-02 (Prompt Correspondence): PASS = 1.0, WARN = 0.5, FAIL = 0.0
\item AB-03 (Create Functionality), AB-04 (View/Edit Operations), AB-05 (Clickable Sweep): PASS = 1.0, WARN = 0.5, FAIL = 0.0
\item AB-06 (Performance): continuous metric normalized to $[0,1]$
\end{itemize}

\begin{table}[!t]
\caption{Check Weights and Definitions Used in Scoring}
\label{tab:check-weights}
\centering
\footnotesize
\begin{threeparttable}
\begin{tabular}{@{}llcp{2.2cm}@{}}
\toprule
\textbf{Check ID} & \textbf{Description} & \textbf{Weight} & \textbf{Notes} \\
\midrule
AB-01 & Boot & 1/6 & Hard gate for $V$ \\
AB-02 & Prompt Correspondence & 1/6 & Hard gate for $V$ \\
AB-03 & Create Functionality & 1/6 &  \\
AB-04 & View/Edit Operations & 1/6 &  \\
AB-05 & Clickable Sweep & 1/6 &  \\
AB-06 & Performance & 1/6 & Normalized to $[0,1]$ \\
\bottomrule
\end{tabular}
\begin{tablenotes}\footnotesize
\item See Section~\ref{sec:scoring} for rubric details. All weights equal after NA re-normalization. AB-01 and AB-02 are hard gates for viability ($V$).
\end{tablenotes}
\end{threeparttable}
\end{table}

\section{Results}
\label{sec:results}

\subsection{Production Deployment and Community Adoption}

The app.build framework has been deployed in production since June 2025, demonstrating real-world viability beyond controlled experiments. The open-source repository (\url{https://github.com/neondatabase/appdotbuild-agent}) has gained significant community traction with 650 stars and 89 forks as of October 2025, indicating strong practitioner interest in environment-first approaches to agentic code generation.

\begin{figure}[!t]
\centering
\includegraphics[width=0.48\textwidth]{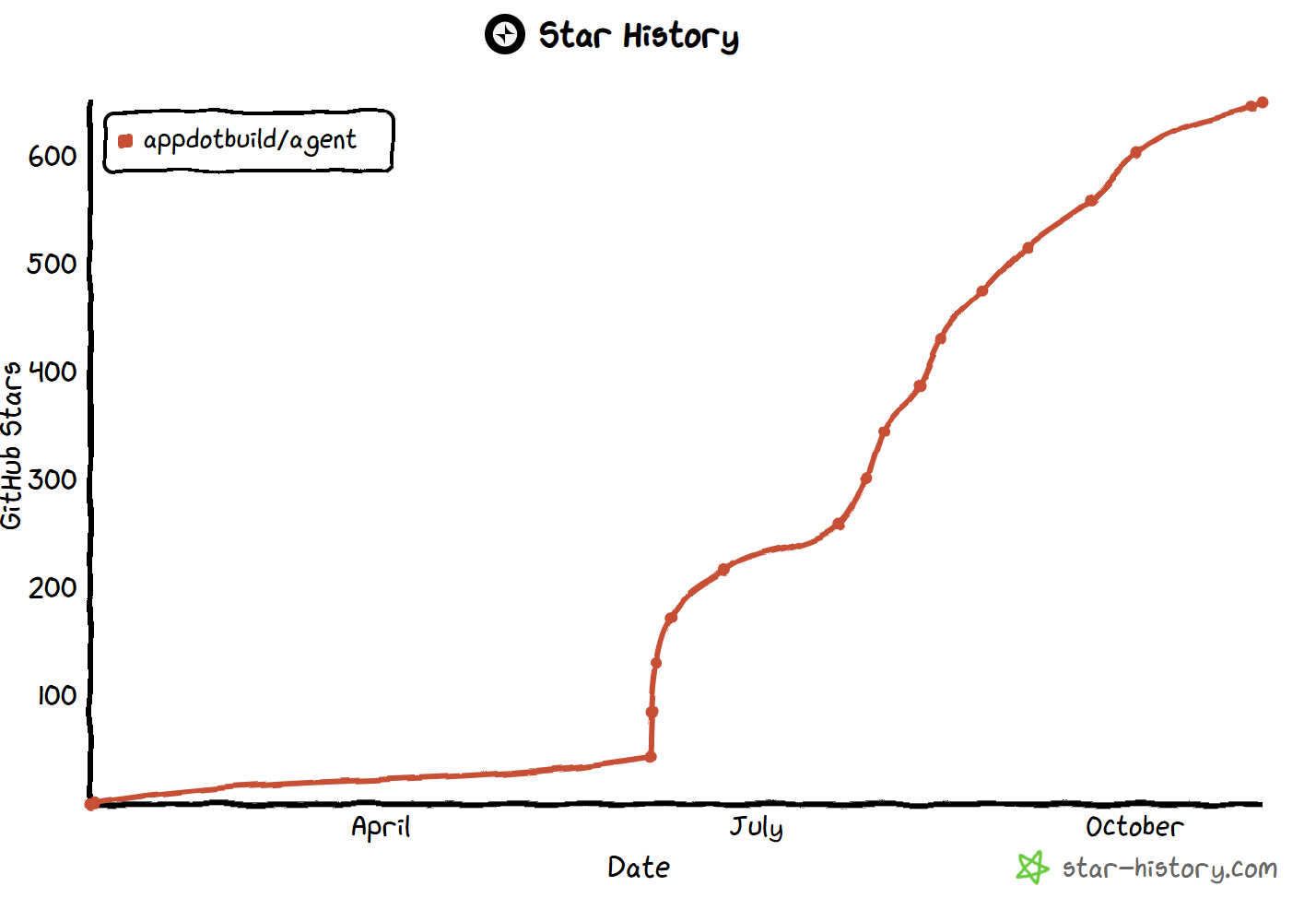}
\caption{GitHub star growth trajectory for appdotbuild/agent repository showing 13x growth over 5 months (May-October 2025), with inflection point in June 2025 coinciding with production deployment launch. The sustained upward trajectory through October 2025 indicates genuine practitioner adoption rather than transient interest. Data from star-history.com.}
\label{fig:star-history}
\end{figure}

\begin{figure}[!t]
\centering
\includegraphics[width=0.48\textwidth]{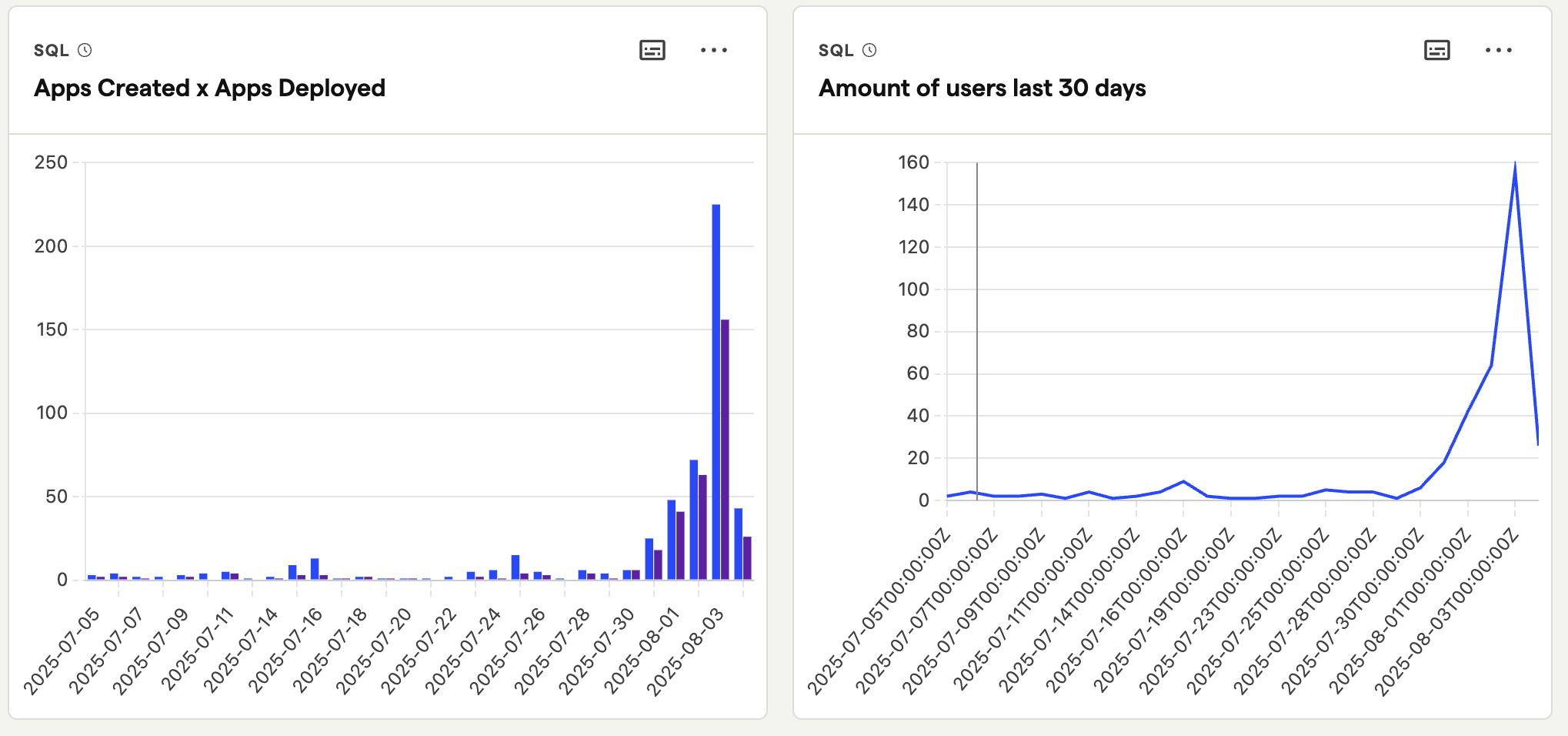}
\caption{Production usage metrics demonstrating real-world deployment scale. Left: Daily application creation and deployment activity showing peak usage of 220+ apps/day in early August 2025. Right: User growth trajectory over 30 days showing rapid adoption spike coinciding with peak usage period, reaching 160+ active users. Data from production database analytics.}
\label{fig:production-usage}
\end{figure}

Figure~\ref{fig:star-history} shows the repository's star growth trajectory, revealing an inflection point in June 2025 when the framework reached production maturity. The repository grew from approximately 50 stars to 650+ stars over five months, representing 13x growth with peak velocity exceeding 100 stars per month during August-September 2025. This organic adoption pattern---characterized by sustained acceleration rather than a single viral spike---suggests the framework addresses genuine practitioner needs.

At peak usage, the platform generated hundreds of applications daily (Figure~\ref{fig:production-usage}, left panel shows peak of 220+ apps/day in early August 2025), with over 3000 user applications generated during the first 4 months (June-October 2025). The concurrent user growth spike (Figure~\ref{fig:production-usage}, right panel) demonstrates sustained platform adoption beyond initial experimentation. This production-scale validation complements our controlled experiments: while our systematic evaluation uses 30 prompts with detailed human assessment and 300 experiments with automated metrics, the production deployment provides ecological validity showing the framework operates reliably in uncontrolled real-world conditions with diverse user requirements.

The community adoption metrics (650+ stars, 89 forks) position app.build among actively-used open-source agent frameworks, demonstrating that practitioners value systematic environment scaffolding for production reliability over model-only approaches. The correlation between production deployment launch (June 2025) and rapid community growth validates the industrial relevance of our environment-first approach.

\subsection{Two-Tier Evaluation Methodology}

Our evaluation combines automated checks with manual viability assessment. We conducted \textbf{300 end-to-end generation experiments} across baseline and ablation conditions, collecting objective metrics (success rate, healthcheck pass rate, cost, duration, token usage) for each run. The automated ``healthcheck'' corresponds to AB-01 (Boot), while full viability assessment requires both AB-01 and AB-02 (Prompt Correspondence) as defined in Section~\ref{sec:scoring}. These automated checks enable consistent measurement and cost-effectiveness analysis across configurations. For viability validation, we performed \textbf{detailed human evaluation on 30 representative prompts} using the AB-check rubric (Section~\ref{sec:scoring}), providing nuanced assessment of functional correctness and production readiness that automated metrics cannot capture.

This two-tier approach reflects industrial practice: automated metrics enable systematic comparison across model architectures and validation strategies, while human evaluation validates that generated applications actually work for their intended purpose. Automated gates filter obviously broken candidates, but human assessors determine whether apps meet production deployment criteria.

\subsection{Automated Validation Results at Scale (n=300)}

Table~\ref{tab:large-scale-automated} presents aggregated results from 300 automated experiments across all conditions. The baseline configuration (Claude Sonnet 4 with full validation) achieved 86.7\% automated success rate at \$110.20 total cost for 30 apps. Open-weights models show cost-performance tradeoffs: Qwen3-Coder-480B achieved 70\% success at \$12.68 total cost, delivering an 8.2x cost reduction per viable app (\$0.61 vs \$5.01), while validation ablations reveal systematic patterns discussed in subsequent sections.

\begin{table}[!t]
\caption{Large-Scale Automated Results Across 300 Experiments}
\label{tab:large-scale-automated}
\centering
\footnotesize
\begin{threeparttable}
\begin{tabular}{@{}lccccc@{}}
\toprule
\textbf{Configuration} & \textbf{n} & \textbf{Success} & \textbf{HC Pass} & \textbf{Cost} & \textbf{Dur.(s)} \\
\midrule
Baseline (Claude) & 30 & 86.7\% & 96.7\% & \$110.20 & 478 \\
No Lint & 30 & 93.3\% & 96.7\% & \$70.49 & 496 \\
No Playwright & 30 & 83.3\% & 93.3\% & \$86.17 & 463 \\
No Tests & 30 & 93.3\% & 100\% & \$71.05 & 373 \\
\midrule
Qwen3-480B & 90 & 70.0\% & 86.7\% & \$12.68 & 629 \\
GPT-OSS-120B & 90 & 30.0\% & 43.3\% & \$4.55 & 628 \\
\bottomrule
\end{tabular}
\begin{tablenotes}\scriptsize
\item Success = automated healthcheck (AB-01) + template validation (AB-02) passed; excludes template-only apps with zero functionality. HC Pass = healthcheck (AB-01) only; includes both functional apps and non-functional templates. Cost = total for cohort. Dur. = mean per-app duration. Open model experiments used simplified validation pipeline (AB-01 + AB-02 only).
\end{tablenotes}
\end{threeparttable}
\end{table}

Key findings from automated metrics: (1) Removing comprehensive validation (no\_lint, no\_tests) increases automated success by +6.7\% but reduces costs by~\$40, suggesting validators catch real issues at measurable expense. (2) Playwright removal has minimal impact on automated success (-3.3\%) while saving \$24, indicating E2E brittleness. (3) Open models achieve viable cost-performance tradeoffs for less critical applications.

\subsection{Cost and Token Usage Analysis}

Detailed telemetry from 300 experiments reveals systematic resource consumption patterns. The baseline configuration (Claude Sonnet 4, full validation) consumed 27.7M input tokens and 1.8M output tokens across 30 apps, averaging 923K input and 60K output tokens per app. This translates to \$3.67 per app at standard API rates (\$3/M input, \$15/M output).

\begin{table}[!t]
\caption{Resource Consumption Breakdown by Configuration}
\label{tab:cost-token-breakdown}
\centering
\footnotesize
\begin{threeparttable}
\begin{tabular}{@{}lrrrr@{}}
\toprule
\textbf{Config} & \textbf{In Tok} & \textbf{Out Tok} & \textbf{Cost} & \textbf{Viable} \\
 & \textbf{/App} & \textbf{/App} & \textbf{/App} & \textbf{Cost} \\
\midrule
Baseline & 923K & 60K & \$3.67 & \$5.01 \\
No Lint & 531K & 50K & \$2.35 & \$2.52 \\
No Playwright & 694K & 53K & \$2.87 & \$3.45 \\
No Tests & 531K & 52K & \$2.37 & \$2.54 \\
\midrule
Qwen3-480B & 728K & 26K & \$0.42 & \$0.61 \\
GPT-OSS-120B & 732K & 26K & \$0.15 & \$0.51 \\
\bottomrule
\end{tabular}
\begin{tablenotes}\scriptsize
\item Tok/App = tokens per application (K = thousands). Viable Cost = cost per viable app (total cost / viable count). Open models via OpenRouter at reduced rates.
\end{tablenotes}
\end{threeparttable}
\end{table}

The cost-per-viable-app metric reveals validation overhead: baseline achieves viability at \$5.01 per app (22/30 viable), while removing unit tests reduces this to \$2.54 (24/30 viable) despite similar per-generation costs. This indicates that comprehensive validation both increases initial costs and filters marginal cases, raising the effective cost per successful outcome.

Open-weights models demonstrate dramatic cost advantages: Qwen3-Coder-480B generates viable apps at \$0.61 each (8.2x cheaper than Claude baseline), though at reduced success rates (70\% vs 86.7\%). For large-scale deployment or less critical applications, this represents a viable engineering tradeoff.

Token efficiency varies by validation configuration: linting and unit tests consume substantial input tokens through multi-round validation cycles (baseline: 923K vs no\_tests: 531K), suggesting that validation rigor directly impacts computational cost. The output token counts remain relatively stable (50K-60K), indicating that validation affects iteration count more than generation verbosity.

\subsection{Detailed Quality Assessment (Human Evaluation, n=30)}

Evaluating 30 TypeScript/tRPC applications, we observe that 73.3\% (22/30) achieved viability ($V=1$), with 30.0\% attaining perfect quality ($Q=10$) and 26.7\% non-viable ($V=0$). Once viability criteria are met, generated applications exhibit consistently high quality.

\begin{table}[!t]
\caption{Aggregated Evaluation Results for TypeScript/tRPC ($n=30$)}
\label{tab:aggregated-results}
\centering
\small
\begin{threeparttable}
\begin{tabular}{@{}lcc@{}}
\toprule
\textbf{Metric} & \textbf{Value} & \textbf{Note} \\
\midrule
Total Apps & 30 & tRPC stack only \\
Viability ($V=1$) & 73.3\% & 22/30 viable \\
Perfect ($Q=10$) & 30.0\% & 9/30 perfect \\
Non-viable ($V=0$) & 26.7\% & 8/30 failed \\
Mean Quality & 8.78 & $V=1$ apps only \\
\bottomrule
\end{tabular}
\begin{tablenotes}\footnotesize
\item Viability $V$ and quality $Q$ defined in Section~\ref{sec:scoring}. Perfect = all checks PASS; non-viable = AB-01 (Boot) or AB-02 (Prompt Correspondence) FAIL.
\end{tablenotes}
\end{threeparttable}
\end{table}

\begin{table}[!t]
\caption{Check-Specific Outcomes Across $n=30$ Tasks}
\label{tab:check-pass-rates}
\centering
\small
\begin{threeparttable}
\begin{tabular}{@{}lcccc@{}}
\toprule
\textbf{Check} & \textbf{Pass} & \textbf{Warn} & \textbf{Fail} & \textbf{NA} \\
\midrule
AB-01 (Boot) & 25 & 2 & 3 & 0 \\
AB-02 (Prompt Correspondence) & 19 & 3 & 5 & 3 \\
AB-03 (Create Functionality) & 22 & 2 & 0 & 6 \\
AB-04 (View/Edit Operations) & 17 & 1 & 1 & 11 \\
AB-05 (Clickable Sweep) & 20 & 4 & 1 & 5 \\
AB-06 (Performance) & 23 & 3 & 0 & 4 \\
\bottomrule
\end{tabular}
\begin{tablenotes}\footnotesize
\item See Section~\ref{sec:scoring} for grading criteria. NA = not applicable for prompt type. Pass rates (excl. NA): AB-01 (Boot): 83.3\%, AB-02 (Prompt Correspondence): 70.4\%, AB-03 (Create Functionality): 91.7\%, AB-04 (View/Edit Operations): 89.5\%, AB-05 (Clickable Sweep): 80.0\%, AB-06 (Performance): 88.5\%.
\end{tablenotes}
\end{threeparttable}
\end{table}

Smoke tests (AB-01, AB-02) determine viability. Among viable applications ($V=1$, $n=22$), quality averaged 8.78 with 77.3\% achieving $Q \geq 9$. Non-viability ($V=0$) arises from smoke test failures or missing artifacts.

\subsection{Open vs Closed Model Performance}

We evaluated Claude Sonnet 4 against two open-weights models using the TypeScript/tRPC stack with simplified validation pipeline (AB-01 Boot + AB-02 Prompt Correspondence only) ensuring the app is bootable and renders correctly. Claude achieved 86.7\% success rate, establishing our closed-model baseline at \$110.20 total cost. Qwen3-Coder-480B-A35B reached 70\% success rate (80.8\% relative performance) while GPT OSS 120B managed only 30\% success rate. Both open models were accessed via OpenRouter, resulting in significantly lower costs: \$12.68 for Qwen3 and \$4.55 for GPT OSS.

The performance gap reveals that environment scaffolding alone cannot eliminate the need for capable foundation models. However, leading open-weights models like Qwen3 demonstrate that structured environments can enable production-viable performance at substantially reduced costs. The 8.2x cost reduction per viable app for 19\% performance loss represents a viable tradeoff for many production scenarios.

\textbf{Template detection reveals inflated success rates.} Our evaluation uncovered a critical issue: many apps that pass healthcheck validation are non-functional "Under Construction" templates with zero functionality. For GPT-OSS-120B, 43.3\% passed AB-01 (Boot), but only 30.0\% passed both AB-01 and AB-02 (Prompt Correspondence), with manual inspection revealing that a substantial portion of bootable apps were generic placeholder scaffolding rather than functional implementations. In contrast, Qwen3-Coder-480B-A35B achieved 86.7\% healthcheck pass and 70.0\% true success, showing better prompt adherence with fewer template generations. \emph{The Success column in Table~\ref{tab:large-scale-automated} reports only viable non-template apps that passed AB-02 validation, while HC Pass includes all bootable apps regardless of functionality}, demonstrating why boot checks alone are insufficient for evaluating code generation quality.

Operational characteristics differed notably between model types. Open models required more validation retries, evidenced by higher LLM call counts (4,359 for Qwen3, 4,922 for GPT OSS vs 3,413 for Claude). AB-01 (Boot) pass rates (86.7\% for Qwen3 vs 96.7\% for Claude) indicate open models generate syntactically correct code but struggle with integration-level correctness, emphasizing the importance of comprehensive validation.

\subsection{Ablation Studies: Impact of Validation Layers}

To understand how each validation layer contributes to application quality, we conducted controlled ablations on the same 30-prompt cohort, systematically removing one validation component while keeping others intact. The baseline configuration (all validation layers active) achieved 73.3\% viability with mean quality score $Q=8.06$.

\textbf{Unit test removal trades quality for apparent viability.} Disabling backend handler tests increased viability to 80.0\% (+6.7 pp) but reduced mean quality to $Q=7.78$ ($-0.28$). This paradox reflects that unit tests catch critical CRUD errors: apps boot successfully without them but fail on data operations. AB-04 (View/Edit Operations) pass rates dropped from 90\% to 60\%, indicating that backend validation prevents functional regressions that smoke tests cannot detect. The increased viability metric reflects false positives: apps that appear viable but contain latent data integrity bugs.

\textbf{Linting removal shows mixed effects with modest gains.} Removing ESLint checks increased viability to 80.0\% (+6.7 pp) and slightly improved quality to $Q=8.25$ (+0.19), suggesting some lint rules may be overly restrictive. However, AB-03 (Create Functionality) dropped 8.3 pp and AB-04 (View/Edit) dropped 7.6 pp, indicating ESLint does catch legitimate structural issues. The net positive quality score indicates that strict linting can reject valid alternative implementations, though the effect size is small.

\textbf{E2E test removal significantly improves outcomes (a negative result).} Removing Playwright tests produced the strongest effect: viability increased to 90.0\% (+16.7 pp) with quality improving to $Q=8.62$ (+0.56). AB-02 (Prompt Correspondence) improved +11.8 pp and AB-05 (Clickable Sweep) improved +5.7 pp. This counterintuitive result indicates that E2E tests introduce more false rejections than they catch real defects. Manual inspection of failed Playwright runs reveals three root causes of brittleness: (1) \emph{Over-specified selectors}: tests hardcode element IDs or CSS classes that models generate variably across runs, causing spurious failures despite functional correctness; (2) \emph{Race conditions}: E2E assertions check UI state before async data fetching completes, rejecting apps that function correctly under human-realistic interaction delays; (3) \emph{False negatives from implementation variance}: models generate semantically equivalent but structurally different UIs (e.g., modal dialogs vs inline forms) that satisfy requirements but fail brittle assertions expecting specific Document Object Model (DOM) structure. This finding suggests that comprehensive E2E suites designed for deterministic codebases are poorly suited to probabilistic code generation, where implementation details vary across generation runs while functional correctness remains stable.

\subsection{Synthesis: Optimal Validation Strategy}

Our ablation results reveal systematic trade-offs between validation rigor and success metrics. Unit/handler tests prove essential for data integrity: removing them increases perceived viability (+6.7 pp) but causes real functional regressions, particularly in AB-04 (View/Edit Operations, $-30$ pp drop). ESLint provides modest value with measurable false positives; the small net quality gain (+0.19) and mixed per-check effects suggest selective application of rules targeting actual errors rather than style preferences. E2E tests currently cause more harm than good, with removal yielding +16.7 pp viability and +0.56 quality improvement, indicating these tests reject too many working applications due to brittleness rather than correctness issues.

These findings suggest a pragmatic validation architecture optimized for probabilistic code generation: \emph{retain} lightweight smoke tests (boot verification and primary route checks) combined with backend unit tests for CRUD operations, which together provide high-confidence quality gates; \emph{refine} static analysis to focus on structural correctness and known anti-patterns while relaxing stylistic rules that reject valid implementations; \emph{replace} comprehensive E2E suites with targeted integration tests covering only critical user paths, avoiding brittle assertions on implementation details. This configuration balances defect detection against false rejection rates. For scenarios where quality requirements dominate cost constraints, comprehensive validation including strict E2E tests remains viable, trading lower automated success rates ($-16.7$ pp) for stronger guarantees about production correctness.

\subsection{Failure Mode Analysis}

We systematically categorize observed failure modes and map them to AB check coverage. \emph{Boot/Load failures} (template placeholders, incomplete artifacts) are reliably caught by AB-01 (Boot) during server startup. \emph{Prompt correspondence failures} (generic templates from generation failures) are detected by AB-02 when human evaluators identify that apps render successfully but implement incorrect functionality. \emph{Content Security Policy (CSP) restrictions} (blocked images or media) are partially caught by AB-05 (Clickable Sweep) through visual inspection, though automated checks miss these unless they break critical flows. \emph{UI interaction defects} (unbound event handlers, non-functional controls) are exposed by AB-05 and AB-03 (Create Functionality) through manual interaction; automated E2E tests theoretically detect these but suffer from brittleness as documented in Section~G. \emph{State/integration defects} (data persistence failures across refresh, broken filters, authentication issues) are caught by AB-04 (View/Edit Operations), though backend unit tests provide only partial coverage and miss client-side state management bugs. Finally, \emph{component misuse} (runtime exceptions from incorrect composition) is detected by AB-01 or AB-05 depending on timing, with TypeScript catching some violations at compile time while others surface only during manual testing.

\textbf{Validation blind spots and proposed upgrades.} Our AB check suite exhibits systematic coverage gaps across five dimensions. First, \emph{accessibility violations} go undetected; apps may function for able-bodied users but fail Web Content Accessibility Guidelines (WCAG) standards. We propose integrating \emph{axe-core} for automated accessibility auditing. Second, \emph{mobile responsiveness} remains unchecked as all validation uses desktop viewports; extending smoke tests to include mobile viewport validation would address this. Third, \emph{performance degradation} under load is not measured; AB-06 assesses only initial load time, missing runtime performance issues. \emph{Lighthouse} performance budgets could provide continuous monitoring. Fourth, \emph{security vulnerabilities} beyond CSP violations (Cross-Site Scripting (XSS), Cross-Site Request Forgery (CSRF), injection attacks) are not detected; static analysis via \emph{semgrep} with OWASP rulesets would provide defense-in-depth. Fifth, \emph{data consistency} in concurrent scenarios and transaction isolation remain untested; implementing multi-client test scenarios would validate state consistency guarantees. These proposed upgrades would strengthen production readiness while maintaining the environment scaffolding paradigm of deterministic, automated validation.

\subsection{Prompt Complexity and Success Rate}
\label{sec:prompt-complexity}

We categorize prompts by complexity: \emph{low} (static/single-page UI), \emph{medium} (single-entity CRUD), \emph{high} (multi-entity workflows with custom logic), and analyze success patterns. Medium-complexity CRUD prompts achieve highest quality ($Q=9$--10), reflecting strong scaffolding alignment with data models and handler patterns. Low-complexity UI prompts prove non-uniformly easy: several failed AB-02 (Prompt Correspondence) by generating generic templates rather than task-specific implementations. High-complexity prompts exhibit lower viability due to interaction wiring and state-consistency issues surfaced by AB-04 (View/Edit Operations) and AB-05 (Clickable Sweep).

\textbf{Trajectory patterns reveal scaffolding trade-offs.} Manual inspection of generation logs exposes distinct model reasoning patterns across complexity levels. For medium-complexity CRUD tasks, environment scaffolding provides effective guidance: structured schema definitions constrain data models to valid patterns, type-safe API contracts prevent interface mismatches, and immediate validator feedback enables targeted repairs within 1--2 iterations. Representative trajectory (\texttt{plant-care-tracker}): schema generation passed immediately; API handlers failed unit tests due to missing foreign key constraints, triggering repair with validator feedback; UI generation succeeded on first attempt using type inference from validated API. This demonstrates the intended scaffolding workflow.

Conversely, for low-complexity UI-only tasks, scaffolding introduces bias: models over-engineer solutions using CRUD templates when simple static pages suffice, causing AB-02 failures. The framework's database-first workflow assumes backend requirements even when unnecessary. Failure example (\texttt{birthday-wish-app}): model generated full API routes and database schema for a static greeting card, achieving technical correctness but failing prompt correspondence. This reveals a structural limitation where strong scaffolding can constrain creativity for tasks outside the target domain.

High-complexity tasks expose scaffolding's context limits. Multi-entity relationships and custom business logic require models to maintain architectural context across multiple validation cycles spanning frontend-backend boundaries. Failures cluster in AB-04 (state management) when models lose track of data flow requirements. Validation feedback successfully catches syntax errors but provides limited architectural guidance, forcing models into trial-and-error that exhausts retry budgets before convergence. This suggests that scaffolding effectiveness degrades as task complexity increases beyond the structured patterns the environment encodes.

\subsection{Threats to Validity \& Limitations}
\label{sec:limitations}

Our current framework is limited to CRUD-oriented data applications, focusing on structured workflows with well-defined input-output expectations. While effective for common web application patterns, it does not yet support complex systems or advanced integrations. The evaluation focused exclusively on the TypeScript/tRPC stack because it is the most mature and well-supported stack in our framework, enabling the most reliable assessment of environment scaffolding effectiveness. We chose web application generation as the evaluation domain because it represents the most widely adopted use case among both closed-source commercial generators and academic end-to-end code generation systems, providing a relevant benchmark for comparison. The validation pipeline, though comprehensive, relies on domain-specific heuristics and expert-defined anti-patterns, which may not generalize to novel or edge-case designs. Additionally, our human evaluation protocol, while rigorous, is poorly scalable and constrained by subjectivity in assessing maintainability and user experience nuances.

\textbf{Benchmark applicability.} We do not evaluate on SWE-bench \citep{jimenez2024swe} or HumanEval \citep{chen2021evaluating} because these benchmarks target fundamentally different tasks: repository-level bug fixing and function-level code completion, respectively. SWE-bench evaluates patch correctness against existing test suites in mature codebases, while our work generates complete applications requiring multi-layered validation (schema correctness, API contracts, UI functionality, integration testing). The evaluation methodology mismatch is structural: existing benchmarks assume deterministic test oracles, whereas greenfield application generation requires human assessment of prompt correspondence, as our template detection findings demonstrate---53.6\% of GPT-OSS-120B outputs passed automated healthchecks but were non-functional placeholders. Our ablation studies provide baseline comparisons that demonstrate environment scaffolding's value: configurations with validation layers disabled show that removing backend tests increases apparent viability (+6.7pp) but causes CRUD correctness to drop 30pp, revealing that comprehensive validation trades automated success metrics for functional correctness.

\subsection{Ethics \& Broader Impact}
\label{sec:broader-impact}

The AI agent boom is accelerating, but real industry deployments often fail silently. Without environment scaffolding, we risk massive overengineering of AI models while ignoring the real bottleneck. App.build represents a shift from model-centric to system-centric AI engineering---a critical step toward scaling reliable agent environments. As practitioners emphasize \citep{babushkin2025machine}, production AI systems only become effective when development integrates not just model performance, but core software engineering principles. By open-sourcing both the framework and evaluation protocol, we provide a reproducible, transparent foundation for building and benchmarking agent environments at scale.

Our results suggest that for CRUD-oriented web applications, structured environment scaffolding complements model capability in achieving production reliability. Through systematic validation, stack-specific orchestration, and iterative repair, app.build demonstrates how probabilistic language models can be guided toward dependable software generation within constrained domains.

Ablations reveal clear trade-offs: removing unit tests increases apparent viability but reduces CRUD correctness; removing linting yields small gains with modest regressions; removing Playwright tests improves outcomes by eliminating flaky UI checks. These results support retaining minimal smoke tests for boot and primary flows, structural checks for UI/code consistency, and scoped E2E tests for critical paths only.

For production-oriented agent systems in structured domains, environment engineering with targeted validation layers offers a complementary path to scaling model capability, providing measurable improvements in reliability while managing cost. As model capabilities continue to advance, the systematic integration of validation and iterative repair remains essential for bridging the gap between probabilistic generation and deterministic production requirements.

\section*{Data and Artifact Availability}

To support reproducibility, we release the following artifacts:
\begin{itemize}
\item \textbf{Framework source code:} Open-source implementation at \url{https://github.com/neondatabase/appdotbuild-agent} (experiments used commit \texttt{e362615} from August 14, 2025)
\item \textbf{Evaluation dataset:} Generated applications, configuration files, and generation logs at \url{https://github.com/keugenek/app.build-publications/tree/main/analysis/dataset} (evaluation conducted August 19-20, 2025)
\item \textbf{Analysis and results:} Experimental results, validation data, and one-command reproduction script (\texttt{run\_analysis\_and\_compare.sh}) at \url{https://github.com/keugenek/app.build-publications/tree/main/analysis/results}
\item \textbf{Evaluation assessor handbook:} Detailed grading criteria, example graded applications, and raw human assessment scores~\citep{appbuild2025handbook}
\item \textbf{Prompt dataset:} Complete set of 30 evaluation prompts with complexity ratings (Appendix~\ref{sec:prompt-dataset})
\item \textbf{Docker environment:} Reproducible sandbox configurations included in the framework repository
\end{itemize}

\section*{Acknowledgments}

This submission is prepared in collaboration between Databricks (app.build team) and THWS University of Applied Sciences W\"urzburg-Schweinfurt (CAIRO). We thank the app.build community for their contributions and feedback which have been invaluable in shaping this work. Special thanks to Databricks executive team for supporting the open-source initiative and providing resources for this research. We also thank David Gomes for advocating for the community-centered vision that guided this project.

\bibliographystyle{plainnat}
\bibliography{references}


\appendix
\section{Prompt Dataset}
\label{sec:prompt-dataset}

This appendix provides the complete specification of evaluation prompts used in our experiments. Table~\ref{tab:prompt-dataset} lists all 30 prompts with their complexity classifications, covering low-complexity UI tasks, medium-complexity single-entity CRUD applications, and high-complexity multi-entity workflows with custom business logic.

\begin{table*}[!htbp]
\caption{Complete Prompt Dataset Used in Evaluation ($n=30$)}
\label{tab:prompt-dataset}
\centering
\footnotesize
\begin{tabular}{@{}p{4cm}p{10cm}p{2cm}@{}}
\toprule
\textbf{ID} & \textbf{Prompt (summary)} & \textbf{Complexity} \\
\midrule
plant-care-tracker & Track plant conditions using moods with custom rule-based logic. No AI/ML/APIs. & Medium \\
roommate-chore-wheel & Randomly assigns chores weekly and tracks completion. & Medium \\
car-maintenance-dashboard & Monitor car maintenance history and upcoming service dates. & Medium \\
city-trip-advisor & Suggest tomorrow's trip viability based on weather forecast API. & High \\
currency-converter & Convert currency amounts using Frankfurter API. & Low \\
book-library-manager & Manage book library with CRUD operations, search, and filters. & Medium \\
wellness-score-tracker & Input health metrics, get daily wellness score with trends. & High \\
event-tracker & Basic event tracker with add, view, delete functionality. & Low \\
daily-pattern-visualizer & Log and visualize daily patterns (sleep, work, social time). & High \\
pantry-inventory-app & Track pantry items, expiry notifications, AI recipe suggestions. & High \\
home-lab-inventory & Catalog home lab infrastructure (hardware, VMs, IP allocations). & High \\
basic-inventory-system & Small business inventory with stock in/out transactions. & Medium \\
pastel-blue-notes-app & Notes app with pastel theme, folders, user accounts. & Medium \\
teacher-question-bank & Question bank with quiz generation and export features. & High \\
beer-counter-app & Single-page beer counter with local storage. & Low \\
plumbing-business-landing-page & Professional landing page for lead generation. & Low \\
kanji-flashcards & Kanji learning with Spaced Repetition System (SRS), progress tracking, JLPT levels. & High \\
bookmark-management-app & Save, tag, organize links with search and sync. & Medium \\
personal-expense-tracker & Log expenses, categories, budgets, spending visualization. & Medium \\
gym-crm & Gym CRM for class reservations with admin interface. & High \\
todo-list-with-mood & To-do list combined with mood tracker. & Medium \\
birthday-wish-app & Static birthday card with message and animation. & Low \\
pc-gaming-niche-site & Budget gaming peripherals review site with CMS. & Medium \\
tennis-enthusiast-platform & Social platform for finding tennis partners. & High \\
engineering-job-board & Niche job board for engineering positions. & High \\
indonesian-inventory-app & Inventory management app in Indonesian language. & Medium \\
habit-tracker-app & Track habits, daily progress, visualize streaks. & Medium \\
recipe-sharing-platform & Community platform for sharing recipes. & High \\
pomodoro-study-timer & Minimalistic Pomodoro timer with session logging. & Low \\
cat-conspiracy-tracker & Humorous app tracking cat suspicious activities. & Low \\
\bottomrule
\end{tabular}
\vspace{2mm}
\footnotesize
\textit{Note.} Dataset details in Section~\ref{sec:prompt-dataset-desc}. Complexity rubric in Section~\ref{sec:prompt-complexity}: Low (static/single-page UI), Medium (single-entity CRUD), High (multi-entity/custom logic).
\end{table*}

\end{document}